%% file: main.tex
\documentclass[letterpaper]{article} 
\usepackage{aaai24}  
\usepackage{times}  
\usepackage{helvet}  
\usepackage{courier}  
\usepackage[hyphens]{url}  
\usepackage{graphicx} 
\usepackage{natbib}  
\usepackage{caption} 
\usepackage{algorithm}
\usepackage{algorithmic}
\usepackage{array}
\usepackage{multirow}
\usepackage{utfsym}
\usepackage{times}
\usepackage{epsfig}
\usepackage{graphicx}
\usepackage{amsmath}
\usepackage{amssymb}
\usepackage{dsfont}
\usepackage{pifont}

\newcommand\blfootnote[1]{%
    \begingroup
    \renewcommand\thefootnote{}\footnote{#1}%
    \addtocounter{footnote}{-1}%
    \endgroup
}

\usepackage{newfloat}
\usepackage{listings}
\DeclareCaptionStyle{ruled}{labelfont=normalfont,labelsep=colon,strut=off} 
\floatstyle{ruled}
\newfloat{listing}{tb}{lst}{}
\floatname{listing}{Listing}
\title{Semi-supervised 3D Object Detection with PatchTeacher and PillarMix}
\author {
    Xiaopei Wu\textsuperscript{\rm 1, 2$\dag$}, 
    Liang Peng\textsuperscript{\rm 1}, 
    Liang Xie\textsuperscript{\rm 1}, 
    Yuenan Hou\textsuperscript{\rm 2}, \\
    Binbin Lin\textsuperscript{\rm 3*},
    Xiaoshui Huang\textsuperscript{\rm 2*},
    Haifeng Liu\textsuperscript{\rm 1},
    Deng Cai\textsuperscript{\rm 1}, 
    Wanli Ouyang\textsuperscript{\rm 2}
\blfootnote{Corresponding author. $^\dag$ This work was done during his internship at Shanghai Artificial Intelligence Laboratory.}
}
\affiliations {
    \textsuperscript{\rm 1}State Key Lab of CAD\&CG, Zhejiang University \\
    \textsuperscript{\rm 2}Shanghai AI Laboratory \\
    \textsuperscript{\rm 3}School of Software Technology, Zhejiang University \\
    \{wuxiaopei, pengliang\}@zju.edu.cn
}
\usepackage{bibentry}

\begin{document}
\maketitle

\begin{abstract}
\input{sections/00-abstract.tex}
\end{abstract}

\section{Introduction}
\input{sections/01-introduction.tex}

\section{Related Work}
\input{sections/02-relatedwork.tex}

\section{Methodology}
\input{sections/03-methodology.tex}

\section{Experiments}
\input{sections/04-experiment.tex}

\section{Conclusion}
\input{sections/05-conclusion.tex}

\clearpage
\section{Acknowledgements}
This work was supported in part by The National Nature Science Foundation of China (Grant Nos: 62273301, 62273302, 62036009, 61936006, 62273303), in part by Ningbo Key R\&D Program (No.2023Z231, 2023Z229), in part by Yongjiang Talent Introduction Programme (Grant No: 2022A-240-G), in part by the Key R\&D Program of Zhejiang Province, China (2023C01135), in part by the National Key R\&D Program of China (NO.2022ZD0160101).

\appendix
\subsection{Analysis for Memory Cost and Training Time}
Table \ref{tab:memory_cost} shows the memory cost and latency of 
PV-RCNN \cite{pv-rcnn} with different voxel sizes.
From the results, we can find that when we use a small voxel size
on the complete scene detection, the memory increases to about 40G, which is 
unaffordable for most GPUs. When using a larger model, such as the multi-frame detector, 
the memory overhead may be even greater. Therefore, dividing the complete scene into patches is necessary. As shown in the last experiment of Table \ref{tab:memory_cost},
When we use PatchTeacher, the memory cost is reduced to an acceptable range.
\vspace{-1mm}
\begin{table}[h]
\centering
    \resizebox{0.47\textwidth}{!}{
        \begin{tabular}{c|c|c|c}
        \hline
         PatchTeacher & Voxel Size (m) & Memory (G) & Latency (ms) \\\hline
         &0.100, 0.100, 0.150  &  11.3 &287 \\ \hline
         &0.050, 0.050, 0.050  &  23.0 &392 \\ \hline
         &0.035, 0.035, 0.035  &  39.9 &468 \\ \hline
\ding{51}&0.035, 0.035, 0.035  &  11.2 &265 \\ \hline
        \end{tabular}
    }
\vspace{-1mm}
\caption{Ablation study on memory cost and latency.}
\vspace{-3mm}
\label{tab:memory_cost}
\end{table}

In our experiment, we divide the complete scene into $4\times4$ patches and 
feed two patches to PatchTeacher in each iteration.
Therefore, the training time should increase $\times8$ times.
To speed up the training, our PatchTeacher only selects those patches that have more than
$\alpha$ ($\alpha$=4 by default) ground-truth boxes. In this way, the amount of patches is 
reduced largely. As shown in Table \ref{tab:training_time}, even training PatchTeacher with the 
same epoch as the baseline, the performance can be improved greatly.
When training PatchTeacher for 60 epochs, the performance is close to the result of training for 120 epochs.
Namely, with only $\times2$ training time, PatchTeacher can provide very strong pseudo labels for the student.
\vspace{-1mm}
\begin{table}[h]
\centering
    \resizebox{0.48\textwidth}{!}{
        \begin{tabular}{c|c|c|ccc}
        \hline
    
         \multirow{2}{*}{PatchTeacher} & \multirow{2}{*}{Epoch} & Training &\multicolumn{3}{c}{3D AP/APH @0.7 (LEVEL 2)} \\
            & & Time & Vehicle & Pedestrian & Cyclist \\ \hline
                  &30  &1h  &49.62/48.99 &45.08/34.69 &39.91/36.87 \\ \hline
        \ding{51} &30  &1h  &54.21/53.66 &51.99/43.95 &44.01/42.76 \\ \hline
        \ding{51} &60  &2h  &55.92/55.39 &56.52/49.07 &48.96/47.49 \\ \hline
        \ding{51} &120 &4h  &56.51/56.00 &57.21/50.18 &49.42/48.05 \\ \hline
        \end{tabular}
    }
\vspace{-1mm}
\caption{Training time under Waymo 5\% protocol. }
\vspace{-5mm}
\label{tab:training_time}
\end{table}

\subsection{Semi-Sampling and PseudoAugment}
\begin{figure}[t]
	\begin{center}
		\setlength{\fboxrule}{0pt}
		\fbox{\includegraphics[width=0.40\textwidth]{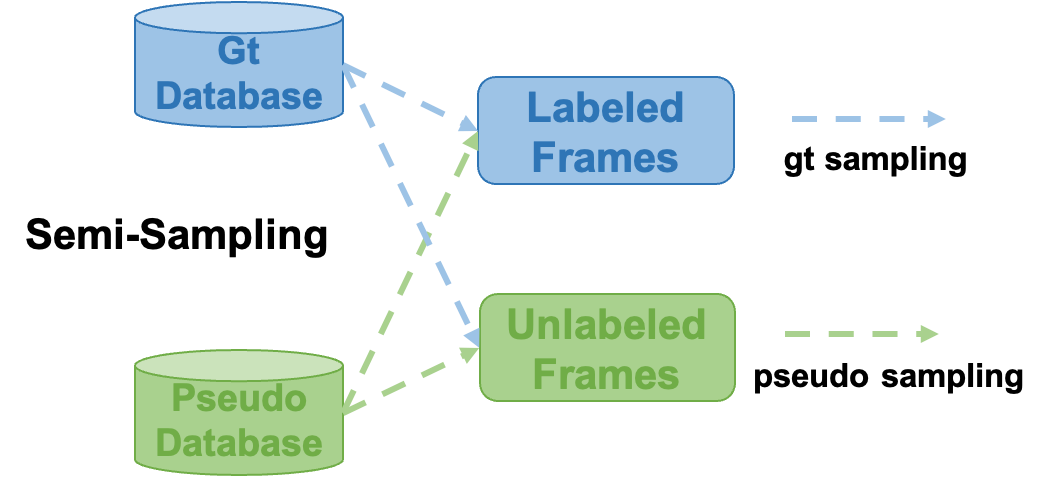}}
	\end{center}
	\vspace{-4mm}
	\caption{Illustration of Semi-Sampling.}
	\label{fig:semi-sampling}
	\vspace{-2mm}
\end{figure}

Gt-sampling \cite{second} is a data augmentation method that uses ground 
truth boxes to crop ground truth samples from labeled frames, and the 
cropped object samples are collected to generate a gt database $\mathcal{G}$.
PseudoAugment \cite{pseudoaugment} leverages pseudo labels of unlabeled 
frames to crop pseudo samples and generate a pseudo database $\mathcal{P}$. It 
pastes samples of $\mathcal{G}$ to unlabeled frames and 
pastes samples of $\mathcal{P}$ to labeled frames.
We unify the above two sampling strategies to Semi-Sampling.
Concretely, Semi-Sampling consists of gt-sampling and pseudo-sampling.
The target frame can be both labeled and unlabeled frames, as shown in 
Figure \ref{fig:semi-sampling}.
Semi-Sampling can be regarded as an improved version of PseudoAugment which adds a pseudo-sampling on unlabeled frames.

Tabled \ref{tab:semi-sampling} shows the effect of each sampling strategy.
Experiment (a) is the baseline which only uses gt-sampling on labeled frames and pseudo-labeling. From experiments (b)-(d), we can find that pseudo-sampling on unlabeled frames is the most effective component. It can be because the amount of pseudo samples and unlabeled frames is huge. Sampling pseudo samples to unlabeled frames can make the most diverse training data. Experiment (e) and (f) is PseduoAugment and Semi-Sampling, respectively. Comparing experiments (e) and (f), we can find that Semi-Sampling is more effective than PseudoAugment due to the pseudo-sampling on unlabeled frames.

\begin{table}[t]
    \begin{center}
    \scalebox{0.65}[0.65]{
        \begin{tabular}{c|cc|cc|ccc}
        \hline
         \multirow{2}{*}{Exp} & \multicolumn{2}{c|}{Labeled Frame} & \multicolumn{2}{c|}{Unlabeled Frame} & \multicolumn{3}{c}{3D AP/APH @0.7 (LEVEL 2)}\\
         \cline{2-8}
        & gt. & pseudo. & gt. & pseudo. & Vehicle & Pedestrian & Cyclist \\ 
        \hline
        (a) & \ding{51} &           & & &53.16/52.60 &49.30/38.26 &44.03/39.83 \\ \hline
        (b) & \ding{51} & \ding{51} & & &54.33/53.77 &48.93/39.71 &44.59/42.70 \\ \hline
        (c) & \ding{51} & & \ding{51} & &53.35/52.78 &49.32/38.62 &45.73/42.19 \\ \hline
        (d) & \ding{51} & & & \ding{51} &54.06/53.42 &50.07/40.19 &48.58/46.51 \\ \hline
        (e) & \ding{51} & \ding{51} & \ding{51} &  &54.16/53.61 &49.08/39.85 &45.94/44.21 \\ \hline
        (f) & \ding{51} & \ding{51} & \ding{51} & \ding{51}  &54.35/53.81 &50.85/40.83 &49.84/47.66 \\ \hline
        \end{tabular}
    }
    \end{center}
    \vspace{-3mm}
    \caption{Ablation of Semi-Sampling. ``gt.'' and ``pseudo.'' represent gt-sampling and pseudo-sampling, respectively.}
    \label{tab:semi-sampling}
    \vspace{-5mm}
\end{table}

\vspace{-1mm}
\subsection{Ablation Study on Score Threshold}
We find that pseudo-labeling (directly use unlabeled frames with pseudo labels for training) and pseudo-sampling need different score threshold 
($\lambda_1$ and $\lambda_2$) for pseudo label filtering. As shown in Table \ref{tab:score_threshold}, when $\lambda_1=0.5$, we can get the best result. With $\lambda_1$ fixed to 0.5, we perform pseudo-sampling and ablate $\lambda_2$. When $\lambda_2=0.8$, we achieve the best result. The reason why $\lambda_2$ is larger than $\lambda_1$ can be that model needs more high-quality pseudo labels for training, and pasting too many low-quality pseudo labels may harm the training.

\vspace{-1mm}
\subsection{Fovea Selection and PillarMix}
Note that PillarMix is not contradictory with Fovea Selection because PillarMix leverages truncated point clouds 
to generate more difficult \textit{\textbf{training}} data, while Fovea Selection avoids truncated point clouds to reduce the difficulty of \textit{\textbf{inference}}.
They all service our SSL framework.

\begin{table}[h]
\vspace{1mm}
\centering
    \resizebox{0.47\textwidth}{!}{
        \begin{tabular}{c|c|cccc}
        \hline
        \multicolumn{2}{c|}{Score} &\multicolumn{4}{c}{3D AP/APH @0.7 (LEVEL 2)} \\
        \multicolumn{2}{c|}{Threshold} & Overall &Vehicle & Pedestrian & Cyclist \\ \hline
        \multirow{5}{*}{$\lambda_1$} & 0.3 &48.46 / 43.31 &53.26 / 52.66	&48.96 / 38.05	&43.17 / 39.22  \\ \cline{3-6}
        & 0.4 &48.80 / 43.50 &53.07 / 52.61	&49.27 / 38.19	&44.07 / 39.70  \\ \cline{3-6}
        & 0.5 &\textbf{48.83 / 43.56} &53.16 / 52.60	&49.30 / 38.26	&44.03 / 39.83  \\ \cline{3-6}
        & 0.6 &47.28 / 42.93 &53.27 / 52.74	&44.74 / 35.75	&43.82 / 40.30  \\ \cline{3-6}
        & 0.7 &44.12 / 40.38 &52.78 / 52.27	&38.66 / 31.05	&40.91 / 37.82  \\ \hline\hline
        \multirow{5}{*}{$\lambda_2$} & 0.5 &49.48 / 45.25 &54.62 / 54.08	&49.66 / 39.74	&44.16 / 41.92  \\ \cline{3-6}
        & 0.6 &50.14 / 45.85 &54.55 / 54.01	&49.74 / 39.71	&46.14 / 43.82  \\ \cline{3-6}
        & 0.7 &50.78 / 46.53 &54.29 / 53.74	&50.37 / 40.23	&47.68 / 45.62  \\ \cline{3-6}
        & 0.8 &\textbf{51.68 / 47.43} &54.35 / 53.81	&50.85 / 40.83	&49.84 / 47.66  \\ \cline{3-6}
        & 0.9 &51.42 / 47.07 &54.33 / 53.80	&51.18 / 40.67	&48.74 / 46.75  \\ \hline
        \end{tabular}
    }
\vspace{-1mm}
\caption{Ablation study on score threshold $\lambda_1$ and $\lambda_2$.}
\vspace{-2mm}
\label{tab:score_threshold}
\end{table}

\begin{figure}[!]
	\begin{center}
		\setlength{\fboxrule}{0pt}
		\fbox{\includegraphics[width=0.42\textwidth]{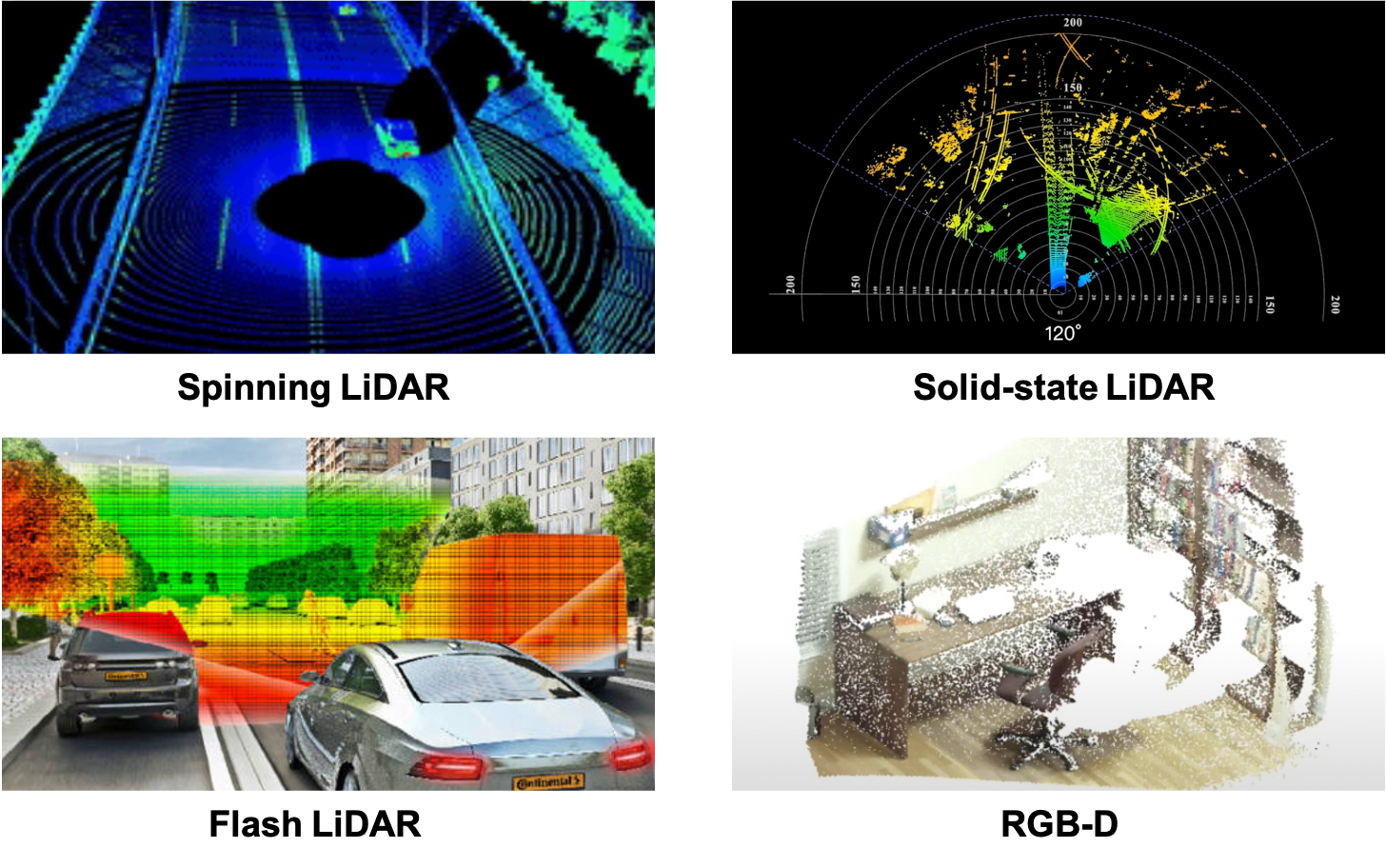}}
	\end{center}
	\vspace{-4mm}
	\caption{Different kinds of point clouds.}
	\label{fig:lidars}
	\vspace{-3mm}
\end{figure}

\subsection{Whether the proposed method can be generalized to non-symmetric point cloud scans?}
The key of \textit{Quadrant Align} is aligning different patches to the first quadrant \textit{to maintain the radiation distribution of point clouds}. By rights, as long as the point clouds are in radiation distribution, 
even if they are not strictly rotationally symmetric, \textit{Quadrant Align} should benefit them.
As shown in Fig \ref{fig:lidars}, although point clouds generated by solid-state LiDAR, flashing LiDAR, or RGB-D cameras are not strictly rotationally symmetric, they are emitted from the origin, making them in radiation distribution.
In addition, considering point clouds generated by solid-state LiDAR, flashing LiDAR, or RGB-D cameras are axisymmetric, 
the \textit{Quadrant Align} can also be designed as a flip instead of a rotation to get rid of the rotation symmetry assumption.

\subsection{Would it be better if two patch teachers are used?}
In Table \ref{tab:distantteacher}, we provide the evaluation of close and far patches for different methods.
The results show that using two PatchTeachers can not achieve better performance.
The reason can be that training patches of different distances separately could decrease the training samples of each teacher.
Under the data-scarce setting, using more training samples could be more important than using expert teachers.

\begin{table}[h]
\centering
    \resizebox{0.48\textwidth}{!}{
        \begin{tabular}{c|c|ccc}
        \hline
    
         \multirow{2}{*}{Method} & \multirow{2}{*}{Distance} &\multicolumn{3}{c}{3D AP/APH @0.7 (LEVEL 2)} \\
        & & Vehicle & Pedestrian & Cyclist \\ \hline
        \multirow{2}{*}{Shared}  & Close &57.52 / 57.01	&57.71 / 50.64	&49.66 / 48.28 \\
        & Far &35.32 / 34.66 &42.67 / 34.71	&29.95 / 27.76 \\ \hline
        \multirow{2}{*}{Not Shared}  & Close &56.55 / 56.09	&56.98 / 48.10	&48.65 / 46.78 \\
        & Far &33.80 / 33.10 &40.89 / 32.09	&24.31 / 21.51 \\ \hline
        \end{tabular}
    }
\vspace{-1mm}
\caption{Evaluation on different distances for PatchTeacher.}
\vspace{-1mm}
\label{tab:distantteacher}
\end{table}

\vspace{-1mm}
\subsection{Qualitative Results}
\begin{figure*}[h]
	\begin{center}
		\setlength{\fboxrule}{0pt}
		\fbox{\includegraphics[width=0.99\textwidth]{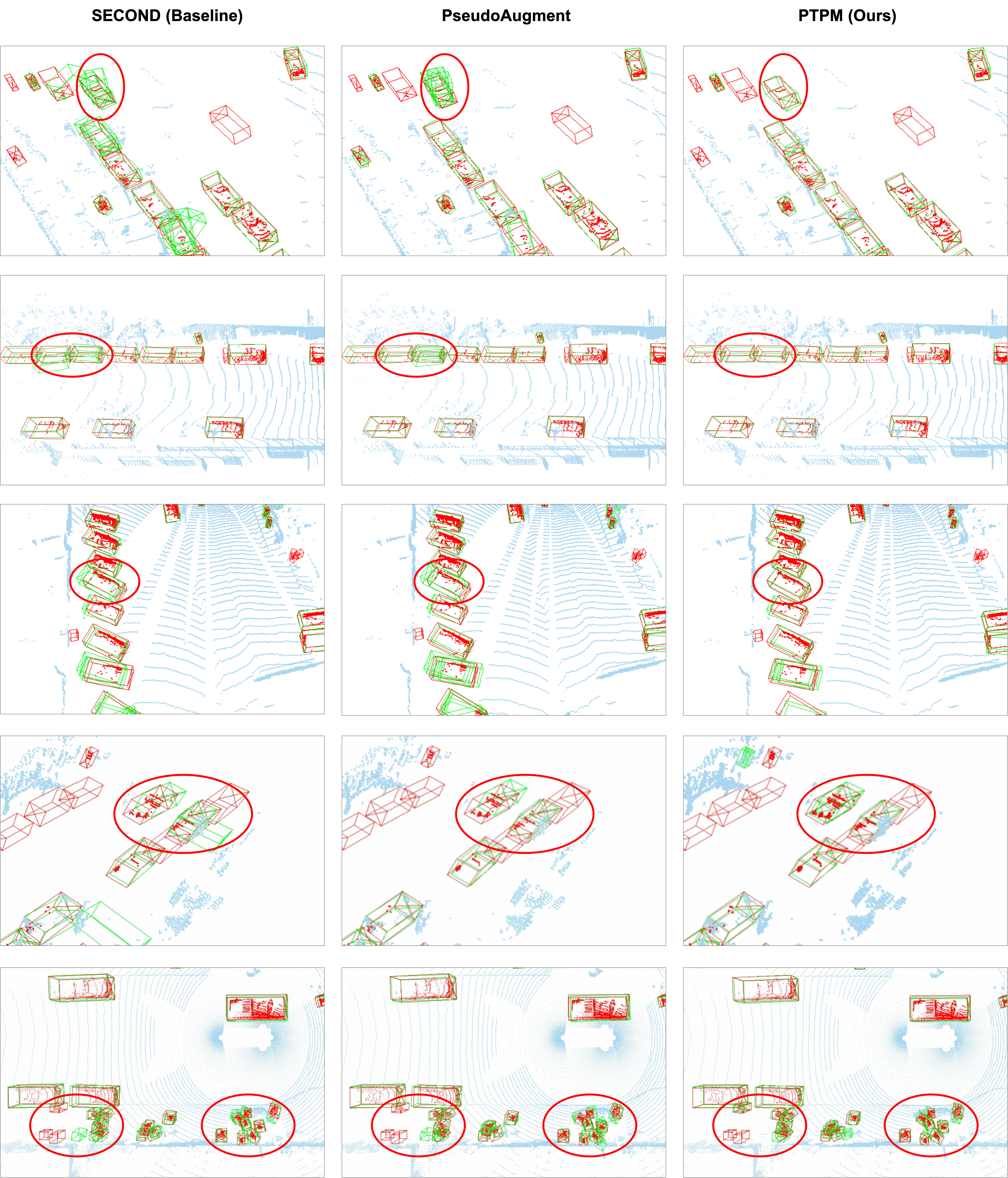}}
	\end{center}
	\vspace{-2mm}
	\caption{Comparison of SECOND, PseduoAugment and our PTPM under Waymo 5\% protocal. We show ground-truth boxes and predictions in red and green, respectively. Points in ground-truth boxes are also rendered in red.}
	\label{fig:vis}
	\vspace{-5mm}
\end{figure*}

We provide the result comparison of different methods as shown in Figure \ref{fig:vis}. Our PTMP can generate more accurate predictions and fewer false positives.

{
\bibliography{aaai24}
}
\end{document}

%% file: sections/00-abstract.tex
Semi-supervised learning aims to leverage numerous unlabeled data to improve the model performance.
Current semi-supervised 3D object detection methods typically use a teacher to generate pseudo labels for a student, and the quality of the pseudo labels is essential for the final performance. In this paper, we propose \textbf{PatchTeacher}, which focuses on partial scene 3D object detection to provide high-quality pseudo labels for the student. 
Specifically, we divide a complete scene into a series of patches and feed them to our PatchTeacher sequentially. PatchTeacher leverages the low memory consumption advantage of partial scene detection to process point clouds with a high-resolution voxelization, which can minimize the information loss of quantization and extract more fine-grained features.
However, it is non-trivial to train a detector on fractions of the scene. Therefore, we introduce three key techniques, i.e., \textit{Patch Normalizer}, \textit{Quadrant Align}, and \textit{Fovea Selection}, to improve the performance of PatchTeacher.
Moreover, we devise \textbf{PillarMix}, a strong data augmentation strategy that mixes truncated pillars from different LiDAR scans to generate diverse training samples and thus help the model learn more general representation.
Extensive experiments conducted on Waymo and ONCE datasets verify the effectiveness and superiority of our method and we achieve new state-of-the-art results, surpassing existing methods by a large margin. Codes are available at \url{https://github.com/LittlePey/PTPM}.

%% file: sections/01-introduction.tex
Recent years have witnessed the rapid development of 3D object detection owing to the boom in deep learning. Many 3D detection methods \cite{pv-rcnn, centerpoint, sfd, mppnet, bevfusion, logonet} have achieved impressive performance on various leaderboards. The appealing performance of these 3D detectors heavily relies on the high-quality annotations of large-scale 3D datasets, which consume enormous human efforts and time. Semi-supervised learning (SSL), which leverages large quantities of low-cost and readily available unlabeled data, is gaining surging attention. 

Current semi-supervised 3D detection methods usually follow the practice of 2D, leveraging an EMA teacher to generate better pseudo labels as the training process proceeds. However, the EMA approach requires the teacher to use the same setting as the student, which limits us from utilizing stronger teachers. 
Recently, some works have noticed this and tried to build a heterogeneous teacher based on or not based on the EMA teacher. For example, \cite{3dal} use a multi-frame teacher to generate accurate pseudo labels for the single-frame student. \cite{proficient-teacher} employs a TTA (test-time augmentation) teacher to supervise the single-forward student. \cite{lpcg} takes advantage of the LiDAR-based teacher to generate strong pseudo labels for the image-based student. However, the aforementioned methods only focus on complete scene detection. The huge amount of points of the complete scene limits their teachers from using high-resolution voxelization to generate more accurate pseudo labels.

To this end, we propose \textbf{PatchTeacher}, which leverages the low memory consumption advantage of partial scene detection to process point clouds with a high-resolution voxelization. It aims to break out the performance bottleneck of the detector trained on the complete scene and produce high-quality pseudo labels for the student. Concretely, we divide a complete scene into $N \times N$ patches and feed them to our PatchTeacher sequentially. Considering that the points of a patch are much fewer than a complete scene, we can use a very small voxel size for voxelization.
The high-resolution voxelization enables the model to extract more fine-grained and discriminative features, which is of great benefit to high-performance 3D object detection.

However, it is non-trivial to train PatchTeacher on point cloud patches.
\textit{Firstly}, distant patches contain few positive samples.
The total loss will be dominated by the negative classification loss of distant patches (see Equation \ref{equ.loss_rpn} for more details and discussions).
Therefore, we propose \textbf{\textit{Patch Normalizer}}.
It normalizes the negative classification loss of different patches in the same frame with the same dominators to prevent the gradient of the classification branch from being dominated by distant patches.
\textit{Secondly}, to train all patches with the same detector, we need to shift them to the same coordinate system, ensuring they are in the same point cloud range.
However, simply shifting patches from different quadrants to the same range will inevitably alter the radiation distribution of point clouds and thus increases the learning difficulty. To tackle this issue, \textbf{\textit{Quadrant Align}} is introduced. It rotates patches from different quadrants to the same quadrant before shifting them to maintain the intrinsic characteristic of LiDAR point clouds.
\textit{Thirdly}, objects located at the edge of patches are difficult to be accurately predicted due to the truncation. We devise \textbf{\textit{Fovea Selection}}, which generates a series of border-overlapped but fovea-non-overlapped patches to address this edge truncated problem. 
Endowed with the above-mentioned strategies, the performance of PatchTeacher is highly boosted.
As a model-agnostic design, our PatchTeacher can empower different students without modifying their architectures and improve semi-supervised learning.

Additionally, given pseudo labels produced by a teacher, how to help the model learn useful information from them is also essential. 
Weak-strong data augmentation is a promising strategy that has been proven in many previous literatures \cite{fixmatch, stac}.
It enforces the model to make consistent predictions for the weakly augmented and the strongly augmented unlabeled data and thus enables the model to learn useful information from the pseudo labels.
Commonly, a stronger data augmentation can help the model learn more information from the pseudo annotations.
Based on this analysis, we propose \textbf{PillarMix}, a strong data augmentation strategy that mixes pillars of different LiDAR scans. 
Specifically, we divide each LiDAR scan into $M\times M$ pillars and then cross-mix these pillars,
ensuring adjacent pillars are always from different LiDAR scans.
This approach produces strongly truncated point clouds on the edges of pillars, which can bring two merits.
First, the blurred edge makes more hard samples which benefits the learning of the model on occluded or sparse samples.
Second, it encourages the model to learn more general features under diverse surrounding environments, which plays a role in regularization.
Experimental results reveal that our PillarMix can achieve impressive improvements in semi-supervised 3D object detection.
To summarize, the contributions of this paper are listed as follows:
\begin{itemize}
    \item We present \textbf{\textit{PatchTeacher}}, which explores the potential of partial scene detection with super high resolution
    to generate strong pseudo labels for semi-supervised 3D detection. In addition, three necessary techniques are devised to improve the performance of PatchTeacher.
    \item We propose \textbf{\textit{PillarMix}}, which cross-mixes pillars from different LiDAR scans to enrich the semi-supervised training data. It enforces the model to make consistent predictions given various incomplete point clouds and surrounding environments.
    \item Our overall pipeline significantly outperforms previous state-of-the-art methods. 
    Extensive experiments on Waymo and ONCE datasets demonstrate
    the effectiveness and superiority of our method.

\end{itemize}

%% file: sections/02-relatedwork.tex
\subsection{Semi-Supervised Learning}
{Semi-Supervised Learning} 
draws growing attention in a wide range of 
research areas.
Since unlabeled data can be obtained more easily than labeled data, 
unlabeled data is far more than labeled data. 
Semi-supervised learning focuses on leveraging the labeled and unlabeled data 
to help the learning of the model. 
Current semi-supervised methods can be roughly divided into pseudo-label-based methods \cite{lee2013pseudo,iscen2019label, arazo2020pseudo} and consistency regularization methods
\cite{bachman2014learning,sajjadi2016regularization, laine2016temporal}.
The former pre-trains a model on labeled data and uses the pre-trained model to infer the unlabeled data.
A predefined threshold is used to filter out the high-quality pseudo labels, and these pseudo labels are
used as the ground truth to re-train the model. 
The latter builds a regularization loss with unlabeled images, which encourages the model 
to generate similar predictions on different perturbations of the same image.

\subsection{Semi-Supervised Object Detection} 
Semi-supervised object detection can also be divided into two groups: consistency-based methods 
\cite{jeong2019consistency, tang2021proposal} and pseudo-labeling methods \cite{stac, unbiased-teacher, soft-teacher}. 
For semi-supervised 3D object detection, SESS \cite{sess} is the pioneering work based on consistency regularization.
\cite{3dioumatch} use the IoU prediction as a localization metric to filter poorly localized proposals.
\cite{3dal, tgnn} utilizes multi-frames to produce more accurate detection results. 
\cite{proficient-teacher} enhances the teacher model to a proficient one with three careful designs.
\cite{detmatch} presents a flexible framework for joint semi-supervised learning on 2D and 3D modalities.
\cite{pseudoaugment} enriches semi-supervised training data with three pseudo-labeling data augmentation policies.
\cite{hssda} devises a dynamic dual-threshold strategy and a shuffle 
data augmentation strategy to improve the performance of SSL.
However, these works focus on complete scene detection, which can not leverage 
high-resolution voxelization to generate more accurate pseudo labels due to memory constrain. In this paper, we leverage partial scene detection to 
build a super high-resolution teacher for high-quality pseudo labels.

%% file: sections/03-methodology.tex
\subsection{Problem Definition}
In semi-supervised 3D object detection, we are given a set of labeled data
$\mathcal{L}={\{L_{i}\}}^{N_{l}}_{i=1}$ and a set of unlabeled data 
$\mathcal{U}={\{U_{i}\}}^{N_{u}}_{i=1}$, where $N_{l}$ and $N_{u}$ denote 
the amount of labeled and unlabeled frames, respectively. 
The objective of semi-supervised 3D object detection is to improve the performance of the model 
with both labeled and unlabeled data.

\subsection{Framework Overview}
\begin{figure*}[t]
	\begin{center}
		\setlength{\fboxrule}{0pt}
		\fbox{\includegraphics[width=1.00\textwidth]{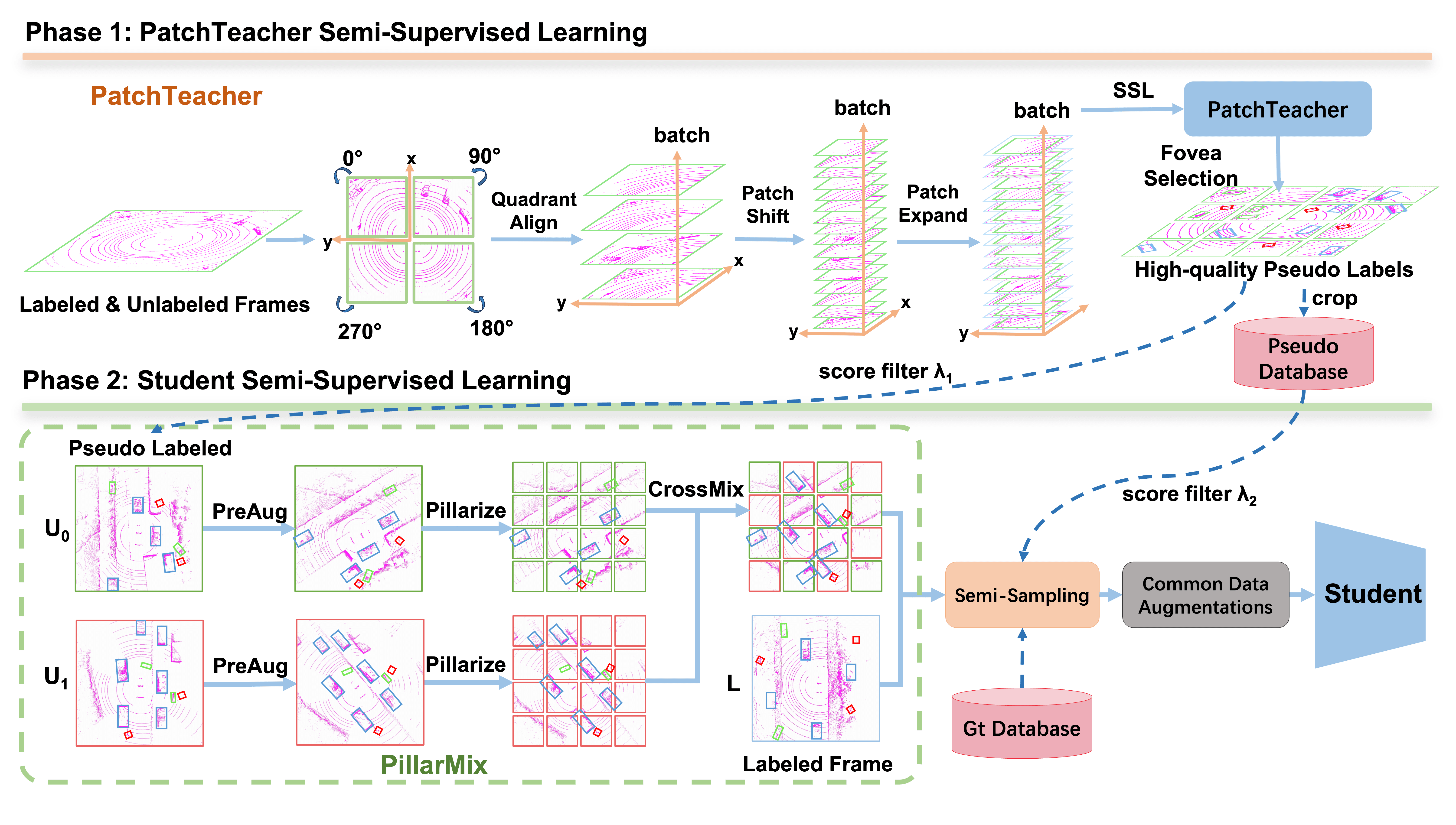}}
	\end{center}
	\vspace{-23pt}
	\caption{Our semi-supervised 3D object detection framework comprises two phases. 
 In phase 1, we train a high-performance PatchTeacher. It focuses on partial scene 
 detection, which enables a super high-resolution voxelization, achieving superior 
 improvement. Three practical techniques and SSL are used to further boost the
 performance of our PatchTeacher. In phase 2, the high-quality pseudo labels produced by 
 PatchTeacher is used to supervise the student model. Given pseudo labels, to make 
 full use of them, we propose PillarMix, which mixes pillars of different LiDAR scans
 crossly, making a strong data augmentation. Then semi-sampling and common data 
 augmentations are followed. Note that semi-sampling is the improved version we develop 
 based on PseudoAugment. The details are provided in the implementation details.}
	\label{fig:overview}
	\vspace{-4mm}
\end{figure*}

As illustrated in Figure \ref{fig:overview}, our semi-supervised framework consists of two phases. In 
phase 1, we train a high-performance PatchTeacher. In phase 2, the student uses the pseudo labels produced by 
PatchTeacher for semi-supervised learning. 
The proposed PillarMix is used to further improve semi-supervised learning.

\subsection{PatchTeacher}
Given a frame of point cloud $\mathcal{P}$, our PatchTeacher uses a divide-and-conquer manner 
to process it. Specifically, we partition $\mathcal{P}$ to $N\times N$ patches, which we denote
${\{p_i\}}^{N\times N}_{i=1}$, and then we process these patches sequentially. 
Each patch contains a small range of point clouds, which only consumes a low memory overhead. This enables us to use an extremely small voxel size to voxelize point clouds. PatchTeacher uses the same architecture as the student while focusing on the
partial scene detection and leveraging high-resolution voxelization to achieve superior performance than the student.
After processing all patches, we merge their predictions to generate the 
final predictions of $\mathcal{P}$. 
However, directly training PatchTeacher can only achieve limited gains.
To this end, we propose three practical techniques to improve the performance of PatchTeacher.

\vspace{2mm}
\noindent\textbf{Patch Normalizer. }
For 3D object detection, the RPN loss is usually formulated as follows:
\begin{equation}\label{equ.loss_rpn}\small
    \mathcal{L}_\text{RPN} = \frac{1}{N_{\text{fg}}}\sum_{i} \mathcal{L}_{\text{cls}}(p_i^a, c_i^{\ast})
    +\frac{1}{N_{\text{fg}}} \mathds{1}(c_i^{\ast}\ge{1})\sum_{i} \mathcal{L}_{\text{reg}}(\delta_i^a, t_i^{\ast}),
\end{equation}
where $N_\text{fg}$ is the number of foreground anchors, which serves as loss normalizer.
$p_i^a$ and $\delta_i^a$ are the outputs of classification and box regression branches, 
$c_i^{\ast}$ and $t_i^{\ast}$ are the classification label and regression targets respectively. 
$\mathds{1}(c_i^{\ast}\ge{1})$ indicates regression loss in only calculated with foreground anchors. 
Due to the occlusion and sparsity properties of point clouds, foreground objects in distant patches are typically rare,
resulting in a small normalizer ($N_\text{fg}$) and a huge amount of easy negatives.
On the contrary,
the foreground objects of nearby patches are usually crowded. Hence, the $N_\text{fg}$ is large and 
there are lots of hard negatives that have a low overlap with the ground truth boxes.
When using loss in Equation \ref{equ.loss_rpn} to train PatchTeacher, gradients of the 
classification branch can be overwhelmed by easy negatives from distant patches whose loss normalizers are small.
This may harm the training and result in poor performance. 

To this end, we propose to normalize the 
classification losses of patches of the same LiDAR scan with the same normalizer, which is motivated 
by the complete scene detection framework. In complete scene detection, the loss of either distant or nearby negatives is normalized by total positive samples.
In high-resolution partial scene detection, due to the memory constraint, we can not feed all patches in a frame to the 
network in an iteration, making it impossible to get the number of total positives. 
Considering the number of positive samples $N_\text{fg}$ and ground-truth boxes are positively correlated, we can multiply the number of the total ground-truth boxes of a scene with a factor $\alpha$ to approximate the number of total positives. 
Formally, we define the Patch Normalizer as $N_\text{p} = \alpha * \sum_{j=1}^{N\times N} G_j$ ($\alpha=3$ by default), 
where $G_j$ is the num of ground truth boxes in the $j^{\mathrm{th}}$ patch. When training PatchTeacher, simply replacing the 
loss normalizer of classification loss from $N_\text{fg}$ to $N_\text{p}$ can obtain good results.

\vspace{2mm}
\noindent\textbf{Quadrant Align. }
To train all patches with the same detector, we need to shift them to the same coordinate system, 
ensuring they are in the same point cloud range.
However, simply shifting patches from different quadrants to the same range will inevitable alter the 
radiation distribution of point clouds which may increase the learning difficulty. 
To alleviate this issue, we align the patches from different quadrants to 
the same quadrant by anticlockwise rotating them around the origin point before shifting.
In this way, the distribution consistency of point clouds in different patches can be guaranteed, making the training more effective.
After \textit{Quadrant Align}, we further divide the point clouds in the same quadrant into several patches evenly. 
To train them with the same model, we use the simple translation operation to
align them to the same point cloud range. We call this \textit{Patch Shift}. 

\begin{figure}[t]
	\vspace{-4pt}
	\begin{center}
		\setlength{\fboxrule}{0pt}
		\fbox{\includegraphics[width=0.44\textwidth]{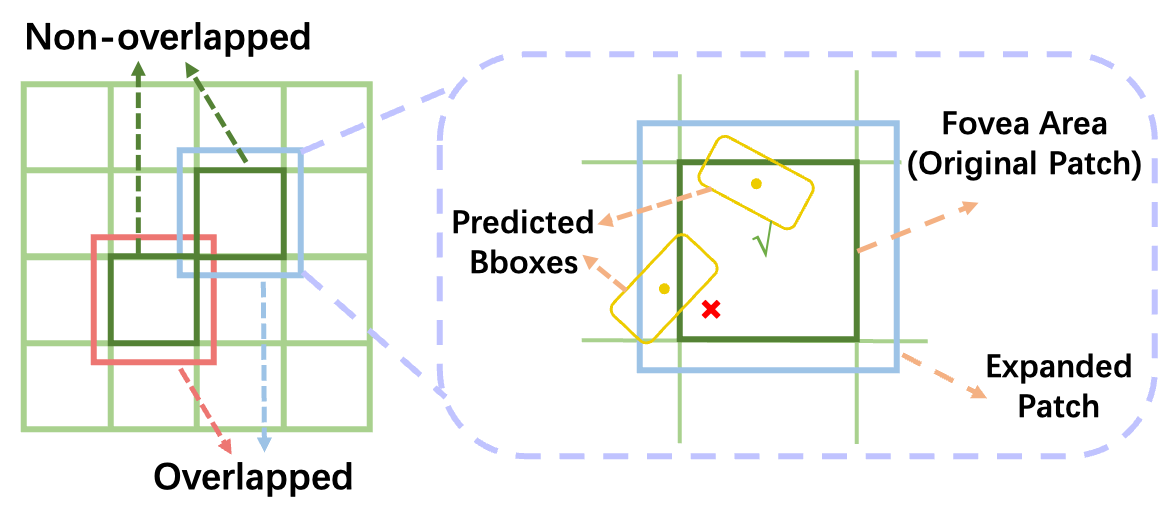}}
	\end{center}
	\vspace{-4mm}
	\caption{Illustration of Fovea Selection. }
	\label{fig:quadrant_fovea}
	\vspace{-5mm}
\end{figure}

\vspace{2mm}
\noindent\textbf{Fovea Selection. }
Considering that the truncated point clouds caused by partition can increase the inference 
difficulty on the edges of the patches, we propose Fovea Selection strategy.
Concretely, we expand the original non-overlapped patches by $\delta$ meters ($\delta$ = 2 by default), 
resulting in a series of overlapped patches, as shown in Figure \ref{fig:quadrant_fovea}. We define the original non-overlapped area of the expanded patch as 
its fovea area. When training PatchTeacher, all points and labels in expanded patches 
will be used. In the inference, the points in expanded patches are also used 
but we only select the predictions whose centers are in the fovea area of expanded patches
to form the predictions of a complete scene. In this way, all predictions we use 
are based on the non-truncated point clouds of expanded patches. This method is especially useful for large objects, such as cars, and Table \ref{tab:patch_teacher} provides the 
ablation study on Fovea Selection.

\vspace{2mm}
\noindent\textbf{Semi-Supervised Learning on PatchTeacher. }
To generate stronger pseudo labels for the student of phase 2, we perform semi-supervised learning 
on the PatchTeacher in phase 1. The semi-supervised learning for PatchTeacher is similar to the student, and the pseudo labels are generated by PatchTeacher itself.

\subsection{PillarMix}

Given two LiDAR scans $U_0$ and $U_1$ and their pseudo labels $P_0$ and $P_1$. 
The output of our PillarMix is a fused LiDAR scan $U_{\mathrm{mix}}$ with its pseudo labels $P_{\mathrm{mix}}$. 
Since the mixing of $P_{\mathrm{mix}}$ and $U_{\mathrm{mix}}$ is similar, here we only describe the mixing of $U_{\mathrm{mix}}$.
To increase the diversity of the fused LiDAR scan, we perform random rotation, random flipping and random scaling 
on $U_0$ and $U_1$ before pillarizing them. Then we divide $U_0$ 
and $U_1$ to $M\times M$ pillars evenly as shown in Figure \ref{fig:overview}. For pseudo label partition, 
a pseudo box belongs to a pillar when its center locates in the pillar. 
Formally, we denote the pillar set of the LiDAR scans $U_i$ as $\mathcal{S}_{i} = \{u_{i}^{jk}, i \in \{0,1\}, j \in [0, M), k \in [0, M)$, $u_{i}^{jk} 
\in \mathbb{R}^{N\times C}\}$ and $u_{i}^{jk}$ denotes the point clouds of the pillar on $j^{th}$ row and 
$k^{th}$ column of the $i^{th}$ LiDAR scan. Considering that the size of foreground objects is small, 
we use a small pillar size to split the scene, which can truncate more foreground objects as more as possible.
With pillars of two LiDAR scans, we can mix them to generate a new LiDAR scan.
An intuitive mixing strategy is randomly selecting some pillars from each LiDAR scan and splicing them. However, it is sub-optimal. 
To make a strong augmentation, we want to leverage each truncated edge of each pillar. Therefore, we alternately select pillars from two LiDAR scans, ensuring adjacent pillars always come from different frames, as shown in Figure \ref{fig:overview}.
Formally, we get a mixed pillar set $\mathcal{S}_{\mathrm{mix}} = \{u_{\mathrm{mix}}^{jk}, j \in [0, M), k \in [0, M)\}$ as follows:
\begin{equation}
    \begin{aligned}
        \mathcal{S}_{\mathrm{mix}} &= \text{PillarMix}(\mathcal{S}_{0}, \mathcal{S}_{1}),\\
    \end{aligned}
\end{equation}
\begin{equation}
    u_{\mathrm{mix}}^{jk}=\left\{
    \begin{array}{ll}
        u_0^{jk}, & j+k \equiv 0 \pmod 2,\\
        u_1^{jk}, & j+k \equiv 1 \pmod 2,\\
    \end{array} \right.
\end{equation}
Then we concatenate $\mathcal{S}_\mathrm{mix}$, resulting in the mixed LiDAR scan $U_\mathrm{mix}$.
Note that the excessive truncation of point clouds can deteriorate the training since the severely truncated objects are difficult to learn and 
could confuse the model. Therefore, the pillar size can not be too small.

\cite{lasermix, polarmix} are LiDAR segmentation data augmentations that fuse laser beams or polars from different LiDAR scans.
Nevertheless, they are incompatible with 3D object detection.
On the one hand, their partition strategy is coarse.
Considering that detection is an object-aware task and the objects are usually small, we need to design a dense partition strategy so that
as many objects as possible can be truncated.
On the other hand, the distance-biased partition strategy they use makes 
the further areas more difficult to truncate and the closer areas to be heavily truncated
, which results in insufficient or excessive data augmentation in these areas. 
Our PillarMix is tailored for 3D object detection, which uses a dense and evenly grid partition strategy, 
augmenting point clouds of various distances strongly, fully and uniformly. 
We provide a detailed ablation study in Table \ref{tab:pillar_comparison} to verify the superiority of our PillarMix over other related methods.

%% file: sections/04-experiment.tex
\subsection{Datasets and Evaluation Metrics}

\vspace{3mm}
\begin{table*}[t]
\centering
\renewcommand\arraystretch{1.00}
    \resizebox{0.92\textwidth}{!}{
        \begin{tabular}{c|c|cccc}
        \hline
        
         \multirow{2}{*}{Label Amounts}  & \multirow{2}{*}{Method} &\multicolumn{4}{c}{3D AP/APH @0.7 (LEVEL 2)}\\
         &  & Overall & Vehicle & Pedestrian & Cyclist \\ \hline

        \multirow{2.5}{*}{~~~5\% ($\sim$ 4k Labels)}  
        &{FixMatch~\cite{fixmatch}}                     & 48.80/43.35    & 51.87/51.27   & 48.28/36.56 & 46.26/42.21 \\  
        &{PseudoAugments\dag~\cite{pseudoaugment}}      & 49.73/45.89    & 54.16/53.61   & 49.08/39.85 & 45.94/44.21 \\
        \multirow{1.0}{*}{~~~$\mathcal{P}^{L}:\mathcal{P}^{U}=1:20$}
        &{ProficientTeacher~\cite{proficient-teacher}}  & 51.10/45.75    & 53.04/52.54   & 50.33/38.67 & 49.92/46.03 \\
        &PTPM (Ours)                                    & \textbf{54.53/51.07}    & \textbf{56.98/56.48}   & \textbf{52.64/44.19} & \textbf{53.96/52.55} \\\hline
        
        \multirow{2.5}{*}{~~~20\% ($\sim$ 16k Labels)}  
        &{FixMatch~\cite{fixmatch}}                     & 55.81/51.45   & 58.94/58.37   & 54.37/44.23   & 54.11/51.75   \\  
        &{PseudoAugments\dag~\cite{pseudoaugment}}      & 55.94/52.29   & 59.55/59.04   & 56.13/47.04   & 52.14/50.79   \\
        \multirow{1.0}{*}{~~~$\mathcal{P}^{L}:\mathcal{P}^{U}=1:5$}
        &{ProficientTeacher~\cite{proficient-teacher}}  & 58.59/54.16   & 59.97/59.36   & 57.88/46.97   & 57.93/56.15   \\
        &PTPM (Ours)                                    & \textbf{60.49/57.02}   & \textbf{61.62/61.16}	& \textbf{59.78/51.17}   & \textbf{60.07/58.74} \\\hline
        
        \multirow{2.5}{*}{~~~100\% ($\sim$ 80k Labels)} 
        &{FixMatch~\cite{fixmatch}}                     & 62.06/57.96   & 63.50/62.98   & 62.00/52.52   & 60.69/58.37 \\
        &{PseudoAugments\dag~\cite{pseudoaugment}}      & 61.15/57.75   & 63.66/63.17	& 61.37/52.74	& 58.42/57.34  \\
        \multirow{1.0}{*}{~~~$\mathcal{P}^{L}:\mathcal{P}^{U}=1:1$}  
        &{ProficientTeacher~\cite{proficient-teacher}}  & 62.96/59.14   & 63.56/63.06   & 62.34/53.19   & 62.97/61.18 \\
        &PTPM (Ours)       &\textbf{65.73/62.13}  & \textbf{67.12/66.6}7	& \textbf{65.85/57.11}	& \textbf{64.23/62.60} \\\hline
        \end{tabular}
    }
\vspace{-1mm}
\caption{Performance on the Waymo Open Dataset with 202 validation sequences. We use the same data split and the same baseline model (SECOND) as ProficientTeacher \cite{proficient-teacher}. PseudoAugments\dag\quad is our implementation.}
\vspace{-1mm}
\label{tab:waymo_second}
\end{table*}

\begin{table*}[h]
\centering
\resizebox{\textwidth}{!}{
    \setlength\tabcolsep{3pt}{\begin{tabular}{c|cc|cc|cc|cc|cc|cc}
        \hline
        \multirow{2}{*}{\begin{tabular}[c]{@{}c@{}}1\% Data \\ ($\sim$ 1.4k scenes)\end{tabular}} & \multicolumn{2}{c|}{Veh. (LEVEL 1)} & \multicolumn{2}{c|}{Veh. (LEVEL 2)} & \multicolumn{2}{c|}{Ped. (LEVEL 1)} & \multicolumn{2}{c|}{Ped. (LEVEL 2)} & \multicolumn{2}{c|}{Cyc. (LEVEL 1)} & \multicolumn{2}{c}{Cyc. (LEVEL 2)} \\
         & mAP & mAPH & mAP & mAPH & mAP & mAPH & mAP & mAPH & mAP & mAPH & mAP & mAPH \\ \hline \hline
        PV-RCNN (from DetMatch) & 47.3 & 45.6 & 43.6 & 42.0 & 28.9 & 15.6 & 26.2 & 14.1 & - & - & - & - \\ \hline
        DetMatch \cite{detmatch} & 52.2 & 51.1 & 48.1 & 47.2 & 39.5 & 18.9 & 35.8 & 17.1 & - & - & - & - \\
        Improvement & +4.9 & +5.5 & +4.5 & +5.2 & +10.6 & +3.3 & +9.6 & +3.0 & - & - & - & - \\ \hline \hline
        PV-RCNN (from HSSDA) & 48.5 & 46.2 & 45.5 & 43.3 & 30.1 & 15.7 & 27.3 & 15.9 & 4.5 & 3.0 & 4.3 & 2.9 \\ \hline
        HSSDA \cite{hssda} & 56.4 & 53.8 & 49.7 & 47.3 & 40.1 & 20.9 & 33.5 & 17.5 & 29.1 & \textbf{20.9} & 27.9 & 20.0 \\
        Improvement & +7.9 & +7.6 & +4.2 & +4.0 & +10.0 & +5.2 & +6.2 & +1.6 & +24.6 & +17.9 & +23.6 & +17.1 \\ \hline \hline
        PV-RCNN (our reproduction) & 47.7 & 38.3 & 44.1 & 33.2 & 27.3 & 14.1 & 22.8 & 11.8 & 5.5 & 4.3 & 5.1 & 3.4 \\ \hline
        PTPM (Ours) &\textbf{61.5}  &\textbf{59.8}  &\textbf{53.7}  &\textbf{52.2}  &\textbf{43.1}  &\textbf{22.3}  &\textbf{36.3}  &\textbf{18.8}  &\textbf{35.7}  &17.9  &\textbf{35.7}  &\textbf{34.3} \\
        Improvement &+13.8  &+21.5  &+9.6  &+19.0  &+15.8  &+8.2  &+13.5  &+7.0  &+30.2  &+13.6  &+30.6  &+30.9 \\\hline
    \end{tabular}
    }
}
\vspace{-2mm}
\caption{Performance comparison on the Waymo Open Dataset with 202 validation sequences for the 3D object detection with PV-RCNN \cite{pv-rcnn} as the baseline model.}
\vspace{-5mm}
\label{tab:waymo_pvrcnn}
\end{table*}

\subsubsection{Waymo Open Dataset}
Waymo~\cite{waymo} is a large-scale LiDAR point cloud dataset, which contains 798 sequences for training and 202 sequences for validation.
We follow ProficientTeachers \cite{proficient-teacher} to divide the 798 training sequences equally into labeled split $\mathcal{P}^{L}$ 
and unlabeled split $\mathcal{P}^{U}$, with each containing 399 sequences. 
Then 5\%, 20\% and 100\% sequences are randomly sampled from $\mathcal{P}^{L}$, leading to the ratio of labeled data and unlabeled data 
$\mathcal{P}^{L}:\mathcal{P}^{U}$ as 1:20, 1:5 and 1:1, respectively. For a fair comparison, we use the same sequence ids 
of 5\%, 20\% and 100\% split as ProficientTeachers \cite{proficient-teacher}. 

\subsubsection{ONCE Dataset} 
ONCE \cite{once} is a large-scale autonomous driving dataset with 1 million LiDAR point cloud samples. 
There are 15k labeled samples, which are divided into 5K for training, 3k for validation and 8k for testing. 
The unlabeled samples are divided into 3 subsets: Small ($\sim$100k samples), Medium ($\sim$500k samples) 
and Large ($\sim$1M samples) to explore the effects of different data amounts for SSL 3D object detection. 

\subsection{Implementation Details}
We use SECOND \cite{second} implemented by OpenPCDet \cite{openpcdet} as our baseline detector, 
following ProficientTeachers \cite{proficient-teacher}. If not specified, the ablation studies 
are conducted under the setting of 5\% labeled Waymo dataset.
For the training of PatchTeacher, each mini-batch consists of 2 patches of labeled point clouds and 2 patches of unlabeled point clouds.
We divide the full point clouds into $4\times 4$ patches. The voxel size of each patch is set to [3.5 cm, 3.5 cm, 3.5 cm].
PatchTeacher is trained for 240 epochs.
For the training of the student, each mini-batch consists of 1 frame of labeled point clouds and 4 frames of unlabeled point clouds.
The student is trained for 30 epochs.
We develop an improved version of PseudoAugment \cite{pseudoaugment}, semi-sampling, to increase the diversity of training data.
Concretely, semi-sampling is a general object sampling strategy, which is a superset of gt-sampling or PseudoAugment. 
Compared to PseudoAugment, semi-sampling adds the feature of pasting samples cropped from an unlabeled frame to another unlabeled frame, 
which is more effective than pasting samples cropped from unlabeled frames to labeled frames or from labeled frames to unlabeled frames. 
More details are provided in the supplementary material.

\vspace{3mm}
\begin{table*}[t]
\label{tab:comparison_3_splits}
\resizebox{0.95\textwidth}{!}{%
    \setlength\tabcolsep{3pt}{\begin{tabular}{cccccccccccccc}
        \hline
        \multicolumn{1}{c|}{\multirow{2}{*}{\textbf{Methods}}} &
          \multicolumn{4}{c|}{\textbf{Vehicle AP (\%)}} &
          \multicolumn{4}{c|}{\textbf{Pedestrian AP (\%)}}  &
          \multicolumn{4}{c|}{\textbf{Cyclist AP (\%)}} &
          \multicolumn{1}{c}{\multirow{2}{*}{\textbf{mAP (\%)}}} \\
        \multicolumn{1}{c|}{} &  overall &  0-30m &  30-50m &
          \multicolumn{1}{l|}{50m-inf} &  overall &  0-30m &  30-50m &
          \multicolumn{1}{l|}{50m-inf} &  overall &  0-30m &  30-50m &  
          \multicolumn{1}{l|}{50m-inf} &  \multicolumn{1}{c}{} \\ \hline
        \multicolumn{1}{c|}{Baseline} &  71.19 &  84.04 &  63.02 &
          \multicolumn{1}{l|}{47.25} &  26.44 &  29.33 &  24.05 &
          \multicolumn{1}{l|}{18.05} &  58.04 &  69.96 &  52.43 &
          \multicolumn{1}{l|}{34.61} &  51.89 \\ \hline
        \multicolumn{14}{c}{\textbf{Small} (100K unlabeled Samples)} \\ \hline
        \multicolumn{1}{c|}{3DIoUMatch} &  73.81 &  84.61 &  68.11 &
          \multicolumn{1}{l|}{54.48} &  30.86 &  35.87 &  25.55 &
          \multicolumn{1}{l|}{18.30} &  56.77 &  68.02 &  51.80 &
          \multicolumn{1}{l|}{35.91} &  53.81\\ 
        \multicolumn{1}{c|}{MeanTeacher} &  74.46 &  86.65 &  68.44 &
          \multicolumn{1}{l|}{53.59} &  30.54 &  34.24 &  26.31 &
          \multicolumn{1}{l|}{20.12} &  61.02 &  72.51 &  55.24 &
          \multicolumn{1}{l|}{39.11} &  55.34\\
        \multicolumn{1}{c|}{ProficientTeacher} & {76.07}  &{86.78}   &{70.19}   &
          \multicolumn{1}{l|}{{56.17}} & {35.90}  & {39.98}  & {31.67}  & 
          \multicolumn{1}{l|}{{24.37}} & {61.19}  &{73.97}   & {55.13}  & 
          \multicolumn{1}{l|}{{36.98}} & {57.72}\\
        \multicolumn{1}{c|}{PTPM (Ours)} & 76.27 & 86.55 & 69.61 &
          \multicolumn{1}{l|}{{56.02}} & 44.29 & 51.95 & 35.86 &
          \multicolumn{1}{l|}{{20.91}} & 61.70 & 75.19 & 54.92 &
          \multicolumn{1}{l|}{{34.57}} & \textbf{60.75}\\
          \hline
          
        \multicolumn{14}{c}{\textbf{Medium} (500K unlabeled Samples)} \\ \hline
        \multicolumn{1}{c|}{3DIoUMatch} &  75.69 &  86.46 &  70.22 &
          \multicolumn{1}{l|}{56.06} &  34.14 &  38.84 &  29.19 &
          \multicolumn{1}{l|}{19.62} &  58.93 &  69.08 &  54.16 &
          \multicolumn{1}{l|}{38.87} &  56.25\\
        \multicolumn{1}{c|}{MeanTeacher} &  76.01 &  86.47 &  70.34 &
          \multicolumn{1}{l|}{55.92} &  35.58 &  40.86 &  30.44 &
          \multicolumn{1}{l|}{19.82} &  63.21 &  74.89 &  56.77 &
          \multicolumn{1}{l|}{40.29} &  58.27\\
        \multicolumn{1}{c|}{ProficientTeacher} &  {78.07} &  {87.43} &{72.50}   & 
          \multicolumn{1}{l|}{{59.51}} &{38.38}  & {42.45}  & {34.62}   & 
          \multicolumn{1}{l|}{{25.58}} &{63.23}  & {74.70} & {58.19}  & 
          \multicolumn{1}{l|}{{40.73}} &{59.89}\\ 
        \multicolumn{1}{c|}{PTPM (Ours)} &76.66 &86.75 &71.30 &
          \multicolumn{1}{l|}{{56.87}} &45.87 &54.98 &37.35 &
          \multicolumn{1}{l|}{{20.89}} &61.88 &74.08 &56.52 &
          \multicolumn{1}{l|}{{33.30}} &\textbf{61.47} \\
          \hline
          
        \multicolumn{14}{c}{\textbf{Large} (1M unlabeled Samples)} \\ \hline
        \multicolumn{1}{c|}{3DIoUMatch} &  75.81 &  86.11 &  71.82 &
          \multicolumn{1}{l|}{57.84} &  35.70 &  40.68 &  30.34 &
          \multicolumn{1}{l|}{21.15} &  59.69 &  70.69 &  54.92 &
          \multicolumn{1}{l|}{39.08} &  57.07\\ 
        \multicolumn{1}{c|}{MeanTeacher} &  76.38 &  86.45 &  70.99 &
          \multicolumn{1}{l|}{57.48} &  35.95 &  41.76 &  29.05 &
          \multicolumn{1}{l|}{18.81} & {65.50} &  75.72 &  60.07 &
          \multicolumn{1}{l|}{43.66} &  59.28\\
        \multicolumn{1}{c|}{ProficientTeacher} & {78.12}  &87.22   &72.74   &
          \multicolumn{1}{l|}{59.58} & {41.95}  &48.09   &35.13   &
          \multicolumn{1}{l|}{26.01} &  {64.12} &75.85   &58.04   &
          \multicolumn{1}{l|}{41.45} & {61.40}  \\
        \multicolumn{1}{c|}{PTPM (Ours)} &76.46 &86.35 &71.31 &
          \multicolumn{1}{l|}{{57.08}} &45.72 &55.00 &36.81 &
          \multicolumn{1}{l|}{{20.25}} &65.87 &77.41 &59.85 &
          \multicolumn{1}{l|}{{42.39}} &\textbf{62.68}\\
          \hline
    \end{tabular}
    }
}
\centering
\vspace{-1mm}
\caption{Evaluations on ONCE validation set with different amounts of unlabeled samples. We compare our PTPM with 3DIoUMatch \cite{3dioumatch}, MeanTeacher \cite{meanteacher} and ProficientTeacher \cite{proficient-teacher}.}
\vspace{-2mm}
\label{tab:once_second}
\end{table*}

\begin{table*}[h]
\centering
\resizebox{0.85\textwidth}{!}{
    \begin{tabular}{c|cccc|cccc}
         \hline
         \multirow{2}{*}{Exp} &\multirow{2}{*}{Pseudo-L} &\multirow{2}{*}{Semi-S} &\multirow{2}{*}{PillarMix}& \multirow{2}{*}{PatchTeacher} & \multicolumn{4}{c}{3D AP/APH @0.7 (LEVEL 2)} \\
         & & & & & Overall & Vehicle & Pedestrian & Cyclist \\ \hline
    
        (a) & & & & & 44.87/40.18 & 49.62/48.99	&45.08/34.69	&39.91/36.87 \\   \hline
        (b) &\ding{51} & & & & 48.63/43.53 & 53.14/52.56	&49.31/38.52	&43.44/39.52 \\   \hline
        (c) &\ding{51} &\ding{51}  & & &51.61/47.43 &54.35/53.81	&50.65/40.83	&49.84/47.66 \\\hline
        (d) &\ding{51} &\ding{51} &\ding{51} & & 52.88/48.75 &55.53/54.98 &51.68/41.93 &51.42/49.33\\\hline
        (e) &\ding{51} &\ding{51} &\ding{51} &\ding{51} & 55.81/52.17 &57.58/57.07	& 54.85/46.23	&55.00/53.20 \\ \hline
    \end{tabular}
}
\vspace{-1mm}
\caption{Ablation study of each component of our method with Waymo 5\% labeled dataset. ``Pseudo-L'' and ``Semi-S'' represent Pseudo-Labeling and Semi-Sampling, respectively.}
\vspace{-5mm}
\label{tab:modules_stack}
\end{table*}

\subsection{Comparison with State-of-the-Arts}
We perform the comparative study for semi-supervised 3D object 
detection based on the SECOND and PV-RCNN on the Waymo dataset.
The results are presented in Table \ref{tab:waymo_second} and  Table \ref{tab:waymo_pvrcnn}.
As depicted in Table \ref{tab:waymo_second}, our method outperforms 
state-of-the-art methods by a large margin under all experimental 
settings of the Waymo dataset. 
Specifically, for the 5\% protocol, we improve mAP from 
ProficientTeacher's 51.10 to 54.53, which achieves 3.43 mAP improvement.
Our PTPM also brings significant improvement when there are more 
labeled data: 58.59 to 60.49 on the 20\% protocol, 
62.96 to 65.73 on the 100\% protocol, 
which demonstrates the effectiveness of our method.
To further evaluate our method, we use PV-RCNN as our baseline 
detector and compare the results with DetMatch \cite{detmatch} and HSSDA \cite{hssda}.
As shown in Table \ref{tab:waymo_pvrcnn}, our method also achieves considerable improvement over
previous methods.
In addition, we compare our method with existing SOTA methods on the ONCE dataset.
The results in Table \ref{tab:once_second} show that our approach also 
exhibits superior results than other methods over all splits. 
Specifically, our method surpasses ProficientTeacer, by 3.03, 1.58, and 1.28 mAP 
on small, medium and large splits, once 
again verifying the advantage of our approaches.
For the ONCE dataset, the improvement to the 
baseline on the pedestrian is the most, which can
be attributed to the high-resolution PatchTeacher.

\subsection{Ablation Study}
To better understand how the proposed approach works, we
conduct a series of ablation studies under the Waymo
5\% labeled data protocol. 

\vspace{1mm}
\subsubsection{Effect of PatchTeacher and PillarMix}
As shown in table \ref{tab:modules_stack}, experiment (a) is the baseline which is only supervised by labeled data.
Experiment (b) uses pseudo-labeling for semi-supervised learning and 
experiment (c) employs the semi-sampling strategy (refer to implementation details) 
to further boost the performance. Comparing experiment (c) with Table \ref{tab:waymo_second}, we can find that only using pseudo-labeling and semi-sampling can achieve better results than ProficientTeacher.
Based on the high-performance experiment (c), our PillarMix further
achieves more than 1 mAP improvement in all classes, as shown in experiment (d).
Moreover, experiment (e) shows that PatchTeacher can yield substantial improvement over experiment (d).

\begin{table}[t]
\vspace{1mm}
\centering
    \resizebox{0.475\textwidth}{!}{
        \setlength\tabcolsep{3pt}{\begin{tabular}{c|cccc|ccc}
        \hline
        \multirow{2}{*}{Exp} &\multirow{2}{*}{P.N.} &\multirow{2}{*}{Q.A.} &\multirow{2}{*}{F.S.} &\multirow{2}{*}{SSL} &\multicolumn{3}{c}{3D AP/APH @0.7 (LEVEL 2)} \\
        & & & & & Vehicle & Pedestrian & Cyclist \\ \hline
        (a) &         &         &         &         &51.83 / 51.26	&54.24 / 45.15	&49.14 / 46.99 \\ \hline
        (b) &\ding{51}&         &         &         &52.61 / 52.09	&54.74 / 45.95	&49.36 / 46.96 \\ \hline
        (c) &\ding{51}&\ding{51}&         &         &55.41 / 54.89	&56.77 / 49.75	&49.58 / 47.59 \\ \hline
        (e) &\ding{51}&\ding{51}&\ding{51}&         &56.51 / 56.00	&57.21 / 50.18	&49.92 / 48.05 \\ \hline
        (d) &\ding{51}&\ding{51}&\ding{51}&\ding{51}&\textbf{58.22 / 57.75}	&\textbf{57.81 / 50.20}	&\textbf{57.61 / 56.20} \\ \hline
        \end{tabular}
        }
    }
\vspace{-1mm}
\caption{Effect of different components of PatchTeacher. ``P.N.'', ``Q.A.'', ``F.S.'' and ``SSL'' represent Patch Normalizer, Quadrant Align, Fovea Selection and Semi-Supervised Learning, respectively. }
\vspace{-6mm}
\label{tab:patch_teacher}
\end{table}

\subsubsection{Effect of each component of PatchTeacher}
Here we explore the effect of three key techniques devised for PatchTeacher.
As shown in Table \ref{tab:patch_teacher}, experiment (a) is
the baseline of PatchTeacher, which uses vanilla partial scene detection.
Experiments (b), (c) and (d) utilize Patch Normalizer, 
Quadrant Align and Fovea Selection, respectively.
From the results, we conclude that each component is 
beneficial for performance, and vehicle and pedestrian 
classes are improved the most. To further improve the
performance, we utilize SSL on our PatchTeacher.
Concretely, we leverage pseudo-labeling, semi-sampling and 
PillarMix to train PatchTeacher like the student.
From the comparison between experiments (d) and (e), we note 
that the SSL contributes the most to the cyclist because the amount of cyclists is small, and its performance is not 
saturated until we leverage a large number of pseudo labels 
of the cyclist in unlabeled data.
Eventually, we improve the simple partial scene detection
by absolutely 6.39, 3.57 and 8.47 mAP on vehicle, pedestrian and cyclist, respectively. The improvement to the complete scene detection (the student) is even more, comparing experiment (d) of Table \ref{tab:patch_teacher} and experiment (a) of Table \ref{tab:modules_stack}. We also evaluate the pseudo labels generated by the student and PatchTeacher on the \textit{unlabeled split}, as shown in Table \ref{tab:teacher_student_comparison}. The results show that our PatchTeacher can achieve more than 10 mAPH improvement over the student, which verifies the rationality of our method.

\vspace{-2mm}
\begin{table}[h]
\centering
    \resizebox{0.475\textwidth}{!}{
        \setlength\tabcolsep{3pt}{\begin{tabular}{c|cccc}
        \hline
         \multirow{2}{*}{Model} &\multicolumn{4}{c}{3D AP/APH @0.7 (LEVEL 2)} \\
        & Overall & Vehicle & Pedestrian & Cyclist \\ \hline
        Student  &46.20 / 41.40 &51.39 / 50.75	&47.11 / 36.06	&40.09 / 37.40 \\ \hline
        Teacher &56.53 / 53.49 &57.74 / 57.25 &60.29 / 53.23 &51.56 / 49.98 \\ \hline
        \end{tabular}
        }
    }
\vspace{-2mm}
\caption{Evaluation results of the student and teacher (PatchTeacher) on the Waymo unlabeled split.}
\vspace{-5mm}
\label{tab:teacher_student_comparison}
\end{table}

\begin{table}[h]
\centering
    \resizebox{0.475\textwidth}{!}{
        \setlength\tabcolsep{3pt}{\begin{tabular}{c|cccc}
        \hline
         \multirow{2}{*}{Voxel (m)} &\multicolumn{4}{c}{3D AP/APH @0.7 (LEVEL 2)} \\
        & Overall & Vehicle & Pedestrian & Cyclist \\ \hline
        0.100  &46.21 / 41.92 &54.30 / 53.71	&43.02 / 33.13	&41.30 / 38.93 \\ \hline
        0.050  &52.85 / 49.92 &55.39 / 55.45	&54.16 / 46.70	&49.01 / 47.60 \\ \hline
        0.040  &53.86 / 50.78 &56.17 / 55.66	&56.41 / 48.94	&49.00 / 47.75 \\ \hline
        0.035  &\textbf{54.38 / 51.41} &56.51 / 56.00	&57.21 / 50.18	&49.42 / 48.05 \\ \hline
        0.003  &54.23 / 51.30 &56.31 / 55.81	&57.10 / 50.07	&49.27 / 48.02 \\ \hline
        \end{tabular}
        }
    }
\vspace{-1mm}
\caption{Ablation study of voxel size of PatchTeacher.}
\vspace{-5mm}
\label{tab:voxel_size}
\end{table}

\begin{table}[h]
\centering
    \resizebox{0.47\textwidth}{!}{
        \setlength\tabcolsep{3pt}{\begin{tabular}{c|cccc}
        \hline
    
         \multirow{2}{*}{$\alpha$} &\multicolumn{4}{c}{3D AP/APH @0.7 (LEVEL 2)} \\
           & Overall & Vehicle & Pedestrian & Cyclist \\ \hline
        1  &57.64 / 54.46 &58.10 / 57.53	&57.74 / 50.04	&57.09 / 55.81 \\ \hline
        2  &57.55 / 54.45 &58.13 / 57.65	&57.73 / 50.12	&56.80 / 55.58 \\ \hline
        3  &\textbf{57.88 / 54.72} &58.22 / 57.75	&57.81 / 50.20	&57.61 / 56.20 \\ \hline
        4  &57.83 / 54.72 &58.35 / 57.89	&57.66 / 50.18	&57.47 / 56.08 \\ \hline
        Avg &48.64 / 46.33 &55.43 / 55.06	&44.83 / 39.18	&45.65 / 44.74 \\ \hline

        \end{tabular}
        }
    }
\vspace{-1mm}
\caption{Ablation study of Patch Normalizer $\alpha$. ``Avg'' represents that we average the negative classification loss by the number of negative samples.}
\vspace{-4mm}
\label{tab:patch_normalizer}
\end{table}

\subsubsection{Effect of hyperparameters}
We now investigate the effect of different hyperparameters in our framework.
We first ablate the voxel size of PatchTeacher. As shown in Table 
\ref{tab:voxel_size}, we observe the best results are achieved when
the voxel size is set to [3.5cm, 3.5cm, 3.5cm], about 1/3 of the 
student that uses a voxel size of [10cm, 10cm, 15cm]. Directly using a small voxel size for voxelization on a complete scene will
increase a huge amount of memory consumption, which is unaffordable.
This emphasizes the importance of partial scene detection.
We then evaluate the effect of Patch Normalizer factor $\alpha$.
The results are shown in Table \ref{tab:patch_normalizer}. When 
$\alpha=3$, the model produces the highest mAP. We also conduct an experiment that 
averages negative classification loss by the number of negative samples. 
As shown in Table \ref{tab:patch_normalizer}, the performance drops drastically.
Then, we explore how the pillar size of PillarMix affects the 
performance of SSL in Table \ref{tab:pillar_size}.
Setting the pillar size too large can not make a strong edge-truncated augmentation, while setting it too small may produce 
over-difficult samples, which is harmful to optimization.
When using a pillar size of 5 meters, the model can achieve the best performance.

\subsubsection{Comparison with PillarMix and other approaches}
To verify the superiority of our PillarMix over other relative approaches, 
we conduct a comparative experiment, as shown in Table \ref{tab:pillar_comparison}.
We apply the LaserMix, PolarMix, Shuffle Data Augmentation and our PillarMix on the baseline, 
which only uses pseudo labeling for SSL.
Here shuffle data augmentation is proposed by \cite{hssda}, which permutes the patches of a LiDAR scan, while it may pull in distant patches 

\begin{table}[h]
\centering
    \resizebox{0.475\textwidth}{!}{
        \setlength\tabcolsep{3pt}{\begin{tabular}{c|cccc}
        \hline
    
         \multirow{2}{*}{Pillar (m)} &\multicolumn{4}{c}{3D AP/APH @0.7 (LEVEL 2)} \\
           & Overall & Vehicle & Pedestrian & Cyclist \\ \hline
        2.5 &52.67 / 48.60 &55.68 / 55.14 &51.67 / 41.88 &50.67 / 48.78 \\ \hline
          5 &\textbf{52.77 / 48.60} &55.44 / 54.87 &51.82 / 42.03 &51.04 / 48.91 \\ \hline
         10 &52.29 / 48.09 &55.16 / 54.59 &51.66 / 41.63 &50.04 / 48.06 \\ \hline
         20 &52.34 / 48.14 &54.83 / 54.27 &51.26 / 41.33 &50.92 / 48.82 \\ \hline
        \end{tabular}
        }
    }
\vspace{-1mm}
\caption{Ablation study of pillar size of PillarMix.}
\vspace{-5mm}
\label{tab:pillar_size}
\end{table}

\begin{table}[h]
\centering
    \resizebox{0.43\textwidth}{!}{
        \setlength\tabcolsep{3pt}{\begin{tabular}{c|ccc}
        \hline
        \multirow{2}{*}{Method} &\multicolumn{3}{c}{3D AP/APH @0.7 (LEVEL 2)} \\
        & Vehicle & Pedestrian & Cyclist \\ \hline
        Baseline &54.35 / 53.81	&50.85 / 40.83	&49.84 / 47.66 \\ \hline
        PolarMix &55.03 / 54.19	&51.05 / 41.33	&50.04 / 48.05 \\ \hline
        LaserMix &55.06 / 54.34	&51.19 / 41.14	&49.39 / 47.37 \\ \hline
        ShuffleDA &54.96 / 54.23	&50.97 / 40.88	&50.19 / 48.00 \\ \hline
        PillarMix &\textbf{55.44 / 54.87}	&\textbf{51.82 / 42.03}	&\textbf{51.04 / 48.91} \\ \hline
        \end{tabular}
        }
    }
\caption{Comparsion of PillarMix and other data augmentation approaches: PolarMix \cite{polarmix}, LaserMix \cite{lasermix} and ShuffleDA \cite{hssda}.}
\label{tab:pillar_comparison}
\vspace{-2mm}
\end{table}

\noindent and push away close point clouds, which destroys the intrinsic characteristic of LiDAR point clouds.
The results show that each approach can improve the baseline, 
while our PillarMix exhibits better results, 
which can be attributed to the sufficient and uniform partition of PillarMix.

\subsubsection{Partial scene detection on different detectors}
Partial scene detection can break the memory 
bottleneck to leverage a extremely small voxel size for feature extraction. 
Here we explore partial scene performances of three mainstream 
detectors.
As shown in Figure \ref{tab:upper_bound}, when using partial scene detection,
we achieve significant improvement over the complete scene detection with different detectors. 

\begin{table}[h]
\centering
    \resizebox{0.47\textwidth}{!}{
        \begin{tabular}{c|c|ccc}
        \hline
         \multirow{2}{*}{Method} & Partial Scene &\multicolumn{3}{c}{3D AP/APH @0.7 (LEVEL 2)} \\
        & Detection & Vehicle & Pedestrian & Cyclist \\ \hline
        \multirow{2}{*}{SECOND}    &  &49.62/48.99	&45.08/34.69	&39.91/36.87 \\
         & \ding{51} &\textbf{56.51/56.00}	&\textbf{57.21/50.18}	&\textbf{49.42/48.05} \\ \hline
        \multirow{2}{*}{PV-RCNN}   &  &5705/56.15	&52.09/29.07	&48.75/44.51 \\
         & \ding{51} &\textbf{59.61/58.98}	&\textbf{60.48/53.07}	&\textbf{55.50/53.76} \\ \hline
        \multirow{2}{*}{CenterPoint} &  &48.21/47.62	&50.10/43.79	&44.36/43.10 \\
         & \ding{51} &\textbf{57.22/56.77}	&\textbf{60.31/54.74}	&\textbf{47.19/46.24} \\ \hline
        \end{tabular}
    }
\caption{Performance of different detectors with partial scene detection with 5\% labeled data (no unlabeled data).}
\label{tab:upper_bound}
\end{table}

%% file: sections/05-conclusion.tex
In this paper, we aim to improve the performance of semi-supervised 
3D object detection. We propose PatchTeacher, which 
leverages the low memory cost advantage of partial scene
detection to equip a high-resolution voxelization.
The PatchTeacher can generate strong pseudo labels for students, which can greatly
improve semi-supervised learning. Moreover, we present a strong data augmentation,
PillarMix, to further boost the performance of our SSL framework. 
Our method sets a new state of the art for semi-supervised 3D object detection in both Waymo and ONCE datasets.